\documentclass{article}

\usepackage{PRIMEarxiv}

\usepackage[utf8]{inputenc} 
\usepackage[T1]{fontenc}    
\usepackage{hyperref}       
\usepackage{url}            
\usepackage{booktabs}       
\usepackage{amsfonts}       
\usepackage{nicefrac}       
\usepackage{microtype}      
\usepackage{lipsum}
\usepackage{fancyhdr}       
\usepackage{graphicx}       
\graphicspath{{media/}}     

\title{Automatic evaluation of scientific abstracts through natural language processing
\thanks{\textit{\underline{Lopes et al.}}: 
\textbf{Lopes, L. G. O., Vieira, T. M. A., Lira, W. W. M..Automatic evaluation of scientific abstracts through natural language processing}} 
}

\author{
  Lucas G. O. Lopes, Thales M. A. Vieira, William W. M. Lira \\
  Universidade Federal de Alagoas\\
  Maceio-AL\\
  \texttt{lucasomena9@gmail.com} \\
  \texttt{thalesv@gmail.com} \\
  \texttt{william@lccv.ufal.br} \\
}

\begin{document}
\maketitle

\begin{abstract}
This work presents a framework to classify and evaluate distinct research abstract texts which are focused on the description of processes and their applications. In this context, this paper proposes natural language processing algorithms to classify, segment and evaluate the results of scientific work. Initially, the proposed framework categorize the abstract texts into according to the problems intended to be solved by employing a text classification approach. Then, the abstract text is segmented into problem description, methodology and results. Finally, the methodology of the abstract is ranked based on the sentiment analysis of its results. The proposed framework allows us to quickly rank the best methods to solve specific problems. To validate the proposed framework, oil production anomaly abstracts were experimented and achieved promising results.
\end{abstract}


\section{Introduction}
The huge amount of research published every day makes challenging to perform literature reviews and track scientific advancements. In this work we propose a framework based on Natural Language Processing (NLP) algorithms to perform several tasks with the aim of evaluating scientific abstracts. The framework is comprised of three main steps:a multi-class text classifier; a recurrent neural network for sequence classification, to segment the abstract; and a sentiment analysis regression to rank the results of the abstracts.
Together, these components are capable of detecting the specific problem of the abstract; segment the abstract text in problem description,  methodology and results; and evaluate the sentiment of the authors of the paper about the results achieved.

We demonstrate the feasibility of our framework through experiments performed on a data set composed of abstracts describing oil production methodologies. In this field, Vargas et al. \cite{VARGAS2019} presented data and studies related to eight different oil production anomalies: 1 - Abrupt Increase of BSW; 2 - Spurious Closure of DHSV; 3 - Severe Slugging; 4 - Flow Instability; 5 - Rapid Productivity Loss; 6 - Quick Restriction in PCK; 7 - Scaling in PCK; 8 - Hydrate in Production Line.

\subsection{Related Works}
The multi-class classifier to identify the texts, related to each abstract problem classes, is developed by studying different approaches: Logistic Regression, Support Vector Machines (SVM), Naive-Bayes classifier, Deep Learning networks combined with Bert \cite{bert} pre-trained embedding layer. Many recent works are focused on deep learning approaches to classify documents. Sousa et al. \cite{romulo2019} proposes a deep network based on bi-Long Short Term Memory (BLSTM) architecture for POS tagging tasks for contemporary and historical Portuguese texts. It uses word embedding and character embedding representations combined with a BLSTM layer. Marulli et al. \cite{Marulli2019} present a similar study for the Italian language, focusing on generalization and to solve the problem of out of the vocabulary (OOV) words. Other recurrent applications include not only topic classification but also topic non-supervised division. Angelov \cite{Angelov}  presents the \textit{top2vec} method for modeling topics, which leverages joint document and word semantic embedding to find $\textit{topic vectors}$. Madrid \cite{Madrid2019} presents a approach to text mining and classification based on meta-learning method. 

The sequence classifier is also a required field in NLP classifiers. Zhiheng et al. \cite{huang2015bidirectional} propose a variety of Long Short-Term Memory (LSTM) based models for sequence tagging. The problem of the text part sequence tagging, in particular, is studied by Dernoncourt et al. \cite{Dernoncourt}. Dernoncourt et al. present different strategies to identify introduction, objectives, methods, results, and conclusions from texts using a labeled corpus from PubMed.

Sentiment analysis is an ever-growing research area, often focused on human interaction and social media. In this context, Peralta \cite{Peralta2019} uses a Markov chain approach to achieve good results on movie review classification.  The work of  Shinha \cite{Shinha2020} used labeled news to determine the sentiment passed by the text: positive, negative, or neutral. The present work uses a similar approach, trying to determine the positive degree of a result of the oil anomaly texts.

In the Natural Language Processing used for oil and gas problems, (NLP) field, Cai et al. \cite{cai2018} show an approach with deep learning-based architectures using NLP methods to predict alarm events in the oil industry, improving real-time decision making.
\section{Methodology}
\begin{figure}[ht]
\centering
\includegraphics[width=0.99\textwidth]{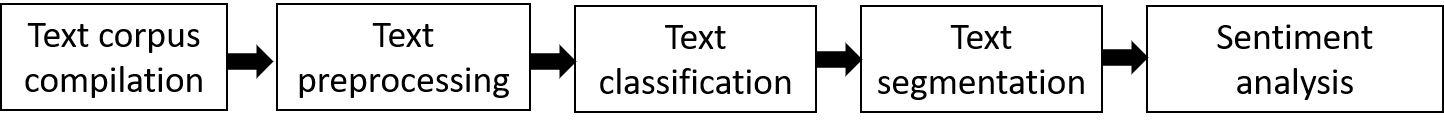}
\caption{Macro-steps for the proposed strategy.}
\label{fig:method}
\end{figure}

The proposed information extraction intends to respond to three questions: 1) what are the methods in the literature to solve these eight unexpected events?; 2) What are the results of these methods?; 3) what are the best methods based on the resulting impact? 
To answer these questions, the present work is divided into four macro-steps, as described in Fig. \ref{fig:method}.

These steps can be better discriminated in the following parts: 1) gather a text corpus; 2) preprocess the text; 3) train an approach to classify the investigated problem using the preprocessed text; 4) segment the text into problem, method or results; 5) rank the abstracts using a sentiment analysis approach of the results segment. The next subsections describe each of these steps. 

\paragraph{Corpus.} This work uses the main corpus, related to the eight problems related to oil production, alongside with two auxiliary corpora, related to text division and sentiment analysis. The main corpus was generated using the OnePetro database of oil industry abstracts and presented by Lopes \cite{lopes2020}. A total of 1600 abstracts, about two hundred for each anomaly, compose the main corpus. The first auxiliary corpus is from the Pubmed database, presented by  Dernoncourt et al. \cite{Dernoncourt}. This corpus presents labels, dividing the abstract texts into background, objectives, methods, results, and conclusion. Shinha \cite{Shinha2020} presents a set of labeled news used as our second auxiliary dataset. The documents are labeled due respect to the sentiment embedded in each sentence as positive, negative, and neutral.

A first analysis of the main corpus is performed using topic modeling and text clusterization. topic modeling process is performed using the Non-negative Matrix factorization (NMF) applied to the Term Frequency–Inverse Document Frequency (TF-IDF) tokenizer. In the attempt to identify the eight subjects of interest, the five more prominent words for each of eight modeled topics are described in the Table \ref{tab:NMFMODEL}: 

\begin{table}[ht]
  \begin{center}
    \caption{Results using TF-IDF and the NMF approach for topic modeling.}
    \label{tab:NMFMODEL}
    \begin{tabular}{c|c} 
      \textbf{Topic} & \textbf{Principal Terms}\\
      \hline
        1 & well, wells, reservoir, gas, sand, oil\\
        2 & gas, hydrates, dissociation, methane, formation, natural\\
        3 & riser, slugging, slug, severe, liquid, pipeline, flow\\
        4 & choke, flow, critical, correlations, chokes, pressure, rate\\
        5 & subsea, system, production, control, design, field, safety\\
        6 & drilling, shale, wellbore, mud, instability, fluid, borehole\\
        7 & scaling, displacement, model, scale, water, oil, flow\\
        8 & fracture, fracturing, proppant, hydraulic, fractures, conductivity, stress\\
    \end{tabular}
  \end{center}
\end{table}

We can see the interposition of the scaling and restriction of the PCK (production choke), that are considered one topic instead of two. Some new topics, such as system and control safety that are not present in the original eight problems, are present in both the tables. The explanation is that all the eight subjects talk, to some extent, about safety issues. The intersection between topics, particularly the restriction and scaling of the PCK, can affect the classification algorithms' performance.

\begin{figure}[h!]
\centering
\includegraphics[width=0.6\textwidth]{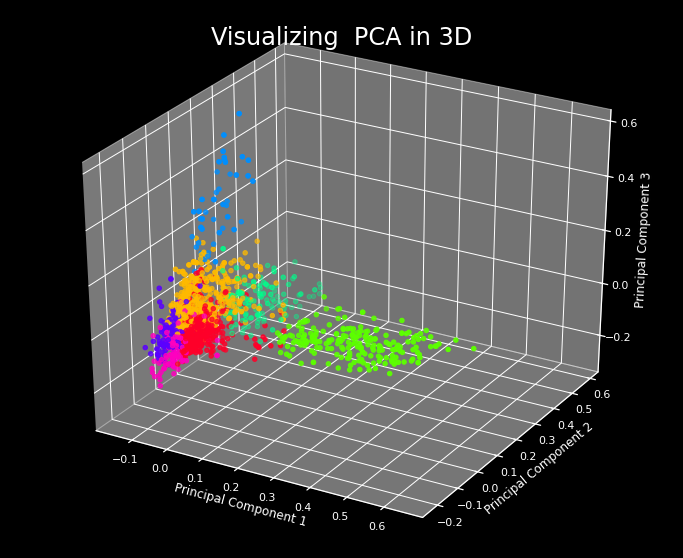}
\caption{PCA decomposition plotting the 3 more prominent components.}
\label{fig:PCA_d}
\end{figure}

We can estimate, using non-supervised text grouping, how many topics our corpus may be divided. Using the Principal Component Analysis (PCA), K-means and the elbow method, we can compare the corpus division's quality. Using the PCA decomposition on the TF-IDF tokens, dividing the corpus into eight principal components give a 98\% of variance preservation. As the value is above 95\%, the number of eight topics can successfully represent the entire corpus. Using the K-means analysis and the elbow method, it is achieved that the optimal K is $7$, which confirms the mixed topic's existence. The topic grouping with the optimal K is evaluated using the PCA approach. The results are described in Fig. \ref{fig:PCA_d} using the three more prominent components of the dimensional decomposition. We can observe that some of the texts are very close to the region of other topics, showing a topic intersection.

\paragraph{Classifying the texts based on the problem} Classification strategies are trained using labeled and processed texts from the eight oil production anomaly cases. 
After the topic modeling, five techniques are tested to classify the anomaly texts: logistic regression, naive-Bayes, support vector machines (SVMs), deep convolutional neural networks, and deep network using the BERT model.

\paragraph{Text segmentation in problem-method-results} Alongside the problem classification, a part of the text classification algorithm is also trained using the Pubmed dataset. This algorithm intends to segment each text on problem, method, and results and is applied in the main corpus after the training. A classifier intended to separate problem-methods-results from an abstract is trained using the PubMed database present in the Dernouncort et al. \cite{Dernoncourt} work. The classifier is trained using a combination of the sentence position and sentence tokens. The position is calculated by the ratio between the number of sentences counted until the respective sentence and the total of sentences in the text. The labels of the PubMed data-set are also adapted: objectives+introduction = problem; methods = methodology; conclusion+results = results.

\paragraph{Rank the methods based on results sentiment analysis} Using Shinha \cite{Shinha2020} dataset and Google Bert deep layer, a single classification algorithm to perform sentiment analysis is trained. This classification is used to rank the part of the texts corresponding to the results. With this evaluation, it is possible to rank the best methods to solve each undesired events.

\section{Results}
This section describes the results of the methods cited in the previous session. Each approach is located in a subsection, showing comparative tables of the different techniques evaluated.

\subsection{Classifying the texts based on the problem}
Different classifiers are tested using labeled texts. Each abstract receives a label corresponding to its respective class (1-8). The classifiers are trained with 70\% of the stratified data, and 30\% of the data are used for validation. The respective classifiers are 1- logistic regression; 2 - Naive-Bayes; 3 - Support Vector Machines (SVM); 4 - convolutional deep network (with embedding layer); 5 - Bert classifier. The first tree approaches are tested using \textbf{Countvectorizer} (CV), \textbf{TF-IDF} and \textbf{Doc2Vec} as input vectorizers to the classifiers. The results of these tree approaches are shown in the Table \ref{tab:fistclass}.  

\begin{table}[h!]
  \begin{center}
    \caption{Quality metrics for different classifiers.}
    \label{tab:fistclass}
    \begin{tabular}{c|c|c|c} 
      \textbf{Classifier} & \textbf{F1-score} &\textbf{Accuracy}&\textbf{Precision}\\
      \hline
        $Logistic Regression (CV)$ &0.712&0.713&0.709\\
        $Logistic Regression (TF-IDF)$ &\textbf{0.722}&\textbf{0.720}&0.725\\
        $Logistic Regression (Doc2Vec)$ &0.348&0.368&0.349\\
        $Naive-Bayes (CV)$ &0.652&0.662&0.669\\
        $Naive-Bayes (TF-IDF)$ &0.601&0.628&0.665\\
        $SVM (CV)$ &0.667&0.662&0.683\\
        $SVM (TF-IDF)$ &0.706&0.697&\textbf{0.735}\\
        $SVM (Doc2Vec)$ &0.354&0.365&0.357\\
    \end{tabular}
  \end{center}
\end{table}

\begin{figure}[h!]
\centering
\includegraphics[width=0.95\textwidth]{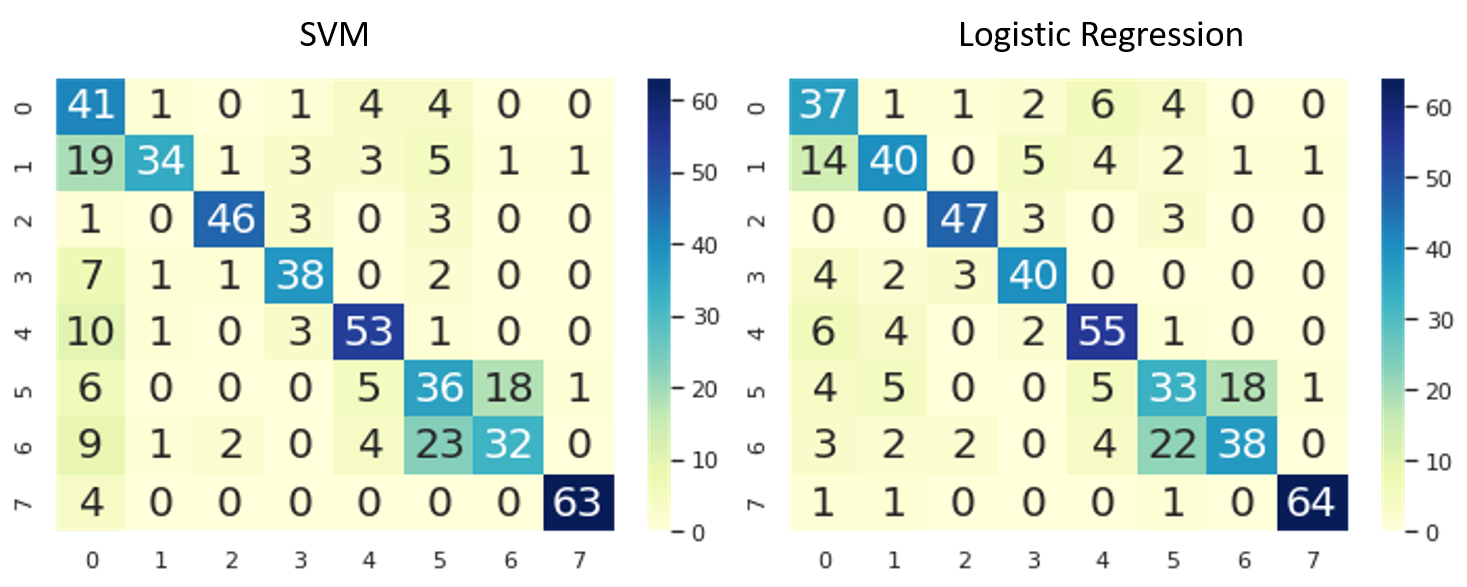}
\caption{Confusion matrix for the SVM and Logistic Regression classifiers.}
\label{fig:logisticsvm}
\end{figure}

\begin{figure}[h!]
\centering
\includegraphics[width=0.95\textwidth]{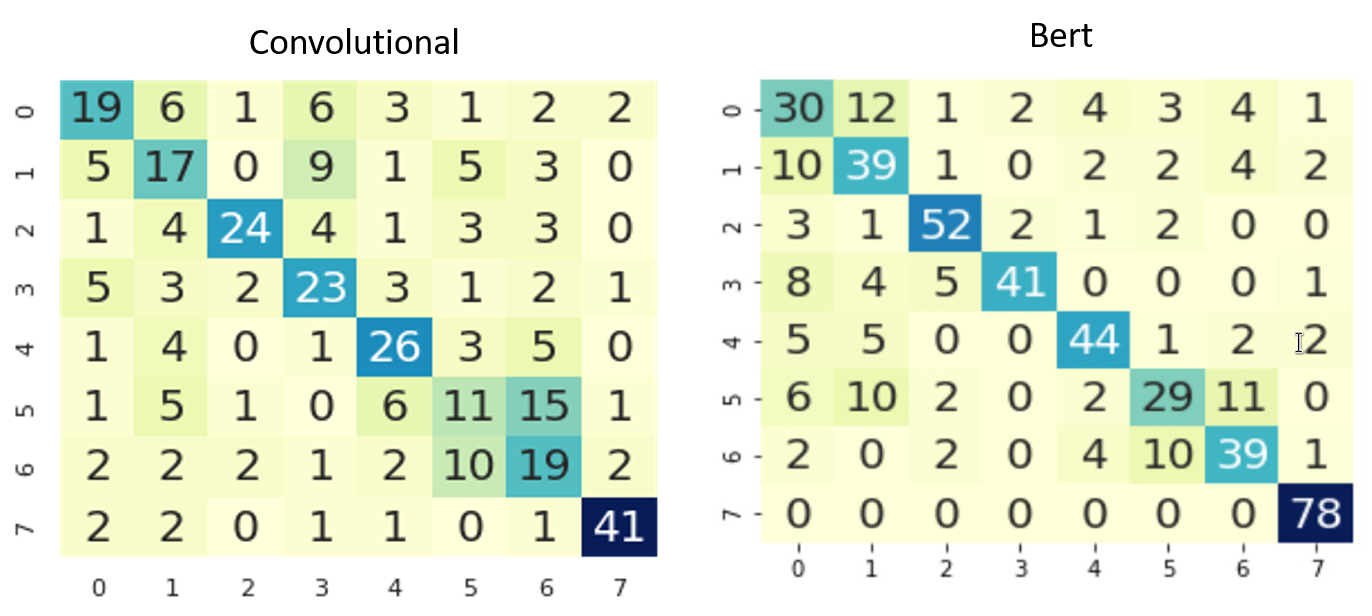}
\caption{Confusion matrix for the deep convolutional and Bert classifiers.}
\label{fig:convolutionalbert}
\end{figure}

We observe that the best values for accuracy and F1-score corresponds to the \textbf{Logistic Regression (TF-IDF)}, and the best precision results are for \textbf{SVM (TF-IDF)}. The confusion matrix for these two approaches are described in the Fig.\ref{fig:logisticsvm}. These confusion matrices show exactly as described in the topic modeling: intersections between classes 7-8 and 1-2. An approach using  embedding and convolutional layers is also described in Table \ref{tab:arquconvbert}. The convolutional network results are described in the confusion matrix in Fig. \ref{fig:convolutionalbert}. This network confusion matrix shows the same limitations due to the class intersection.

\begin{table}[h!]
  \begin{center}
    \caption{Architectures Convotutional and Bert (TF-IDF tokenizer)}
    \label{tab:arquconvbert}
    \begin{tabular}{c|c|c} 
      \textbf{Convolutional} & \multicolumn{2}{c}{\textbf{Bert}}\\
      \hline
        Layer/Output Shape & \multicolumn{2}{c}{Layer/Output Shape}\\
        $(Embedding)$ - (None, 100, 100) & \multicolumn{2}{c}{$(input-word-ids )$ - (None, 72)}\\
        $(Conv1D)$ - (None, 96,128) & \multicolumn{2}{c}{$(tf-bert-model )$ - (None, 72,1024)}\\
        $(global-max-pooling1d )$ - (None, 128) & \multicolumn{2}{c}{$(tf-strided-slice )$ - (None, 1024)}\\
        $(Dropout)$  -  (None, 128) & \multicolumn{2}{c}{$(Dropout)$ - (None, 1024)}\\
        $(Dense)$  -  (None, 4096) & \multicolumn{2}{c}{$(Dense)$  - (None, 128)}\\
        $(Dense)$  - (None, 8) & \multicolumn{2}{c}{$(Dense)$  -(None, 8)}\\
        \hline
        \textbf{Score}  & \textbf{Full-text}  & \textbf{Nouns-text}\\
        F1/Acc/Precision & F1/Acc/Precision  & F1/Acc/Precision \\
        0.543/0.549/0.552 & 0.585/0.598/0.582 & 0.685/0.689/0.691\\
    \end{tabular}
  \end{center}
\end{table}

The last models used are Bert embedding models with a dense layers. This approach used two different versions of the texts: the full abstracts and the only nouns abstracts. We use the Spacy package to extract the text entities and remove the other POS different from a noun. The results using this approach for the other classifiers did not give a better answer than using the full text. However, the Bert classifier improves the prediction scores on the only nouns texts. The results of this approach are presented in Table \ref{tab:arquconvbert}. The results of this approach, using the Nouns training set, are better than the other Bert approaches, and is presented in the Fig. \ref{fig:convolutionalbert}. The results using only nouns are close to the SVM and Logistic Regression.

\subsection{Text segmentation in problem-method-results} \label{sec:division}

 The architecture used for the classifier is based on Long Short-Term Memory architecture, and the used tokenizer is the mean Word2Vec model from Google-News. The network architecture described in Table \ref{tab:arqlstm2}.

\begin{table}[h!]
  \begin{center}
    \caption{Text part classifier based on LSTM archtecture.}
    \label{tab:arqlstm2}
    \begin{tabular}{c|c|c} 
      \textbf{Layer} & \multicolumn{2}{c}{\textbf{Output Shape}}\\
      \hline
        $(LSTM)$ & \multicolumn{2}{c}{(None, 128,300)}\\
        $(Dense)$  & \multicolumn{2}{c}{(None, 32)}\\
        $(Dropout 0.3)$  & \multicolumn{2}{c}{(None, 32)}\\
        $(Dense)$  & \multicolumn{2}{c}{(None, 3)}\\
        \hline
        \textbf{Score} & \textbf{PubMed-texts} & \textbf{Oil Anomaly-texts}\\
        F1-Score/Accuracy/Precision & 0.826/0.824/0.841 & 0.765/0.774/0.769\\
    \end{tabular}
  \end{center}
\end{table}

\begin{figure}[h!]
\centering
\includegraphics[width=0.6\textwidth]{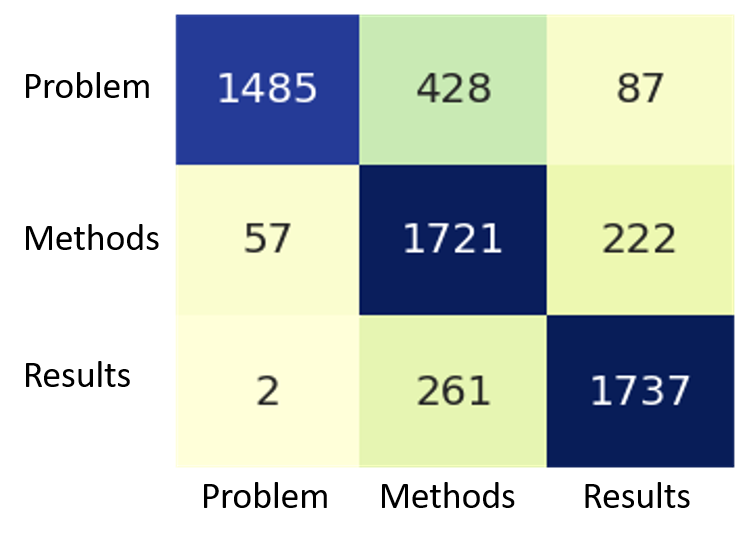}
\caption{LSTM text segmentation confusion matrix.}
\label{fig:pubmed}
\end{figure}

Dernouncort et al. show different approaches to separate the texts obtaining accuracy over 80\% and up to 90\%. The results of Table \ref{tab:arqlstm2} show close results, and add a new position input. We believe that the position input makes the network more general to texts different from the trained corpus. The confusion matrix for the text divider classifier is present in Fig. \ref{fig:pubmed}.
\begin{table}[!t]
  \begin{center}
    \caption{Text part classifier results.}
    \label{tab:sepresults}
    \begin{tabular}{l|l} 
      \textbf{Text} & \textbf{Class}\\
      \hline
        In addition, some wells presented sand production problems. & result\\
        These moduli are, however, difficult to measure in experiments. & result\\
        36.8Ma and then keep vibration in the last ca.36.8Ma. & result\\
        The value of this pressure is called the closure stress. & method\\
        After a peak oil production period, the GOR starts to increase (Ref. & result\\
        Also the fluid flow is considered as a transient flow. & problem\\
        Experimental Procedures Sample Selection and Preparation. & result\\
    \end{tabular}
  \end{center}
\end{table}

The results of this text divider classifier on the oil anomaly data-set is also present in Table \ref{tab:arqlstm2}. These results are achieved by mechanical observation of a stratified set in the oil anomaly text database. The values are indeed inferior to those obtained on the PubMed test set, but still good results. Table \ref{tab:sepresults}
shows the results for text separation.



\subsection{Rank the methods based on results sentiment analysis}
The second auxiliary data-set from Shinha \cite{Shinha2020} is used to train a neural network for sentiment analysis with Bert. The based architecture for a binary classifier (positive or negative) is similar to that presented in Table \ref{tab:arquconvbert} for the Bert model, changing the number of neurons in the last dense layer from 8 to 2.


The results show a good correspondence with both the validation set from Shinha, and the mechanical analysis in the stratified test set using a subset from the validation data. The impact analysis is performed using the sentiment analysis in the results obtained from Section \ref{sec:division}. The bellow sentences show text result's fragments and the respective sentiment analysis on a range from "0" (most negative) to "1" (most optimistic):

"\textit{They concluded that hydrate grows preferentially in coarse-grained sediments because lower capillary pressures in these sediments permit the migration of gas and nucleation of hydrate.The growth of gas hydrate in clay-rich sediments, \textbf{however}, is \textbf{more poorly understood} and appears to be \textbf{limited} to mostly massive occurrences.}" -
\textit{Calculated impact}: \textbf{0.22}

The \textbf{bold} marked words are probably the cause of the lower impact value from sentiment analysis. The following result fragment presents a higher impact value:

"\textit{The application of extended subsea gathering networks and transportation of unprocessed wellstreams are \textbf{amongst favourable options} for \textbf{reducing} field development and operational \textbf{costs}.These lines will convey a cocktail of multiphase fluids, including mixed electrolyte produced water, and liquid and gaseous hydrocarbons. \textbf{A good knowledge} of the behaviour of these complex systems is essential for \textbf{confident and economical design} and operation of associated fields, pipelines and processing facilities.}" -
\textit{Calculated impact}: \textbf{0.99}

The \textbf{bold} marked words are probably the cause of the higher impact value from sentiment analysis. The following text shows an inconclusive analysis:

"\textit{This probabilistic method has been applied to log data acquired from the ODP/IODP (marine) and the Mallik (permafrost) exploration wells, \textbf{successfully generating} gas hydrate saturation estimates, with comparable results to Archie saturation estimates at high saturations. \textbf{Further investigation is required} to fully determine the potential of probabilistic methods for gas hydrate evaluation.}" -
\textit{Calculated impact}: \textbf{0.501}

The results show that the model can adapt to various texts depending on the used words. However, some combinations of words can bring undesired results to the sentiment analysis.

\section{Conclusion}
This work presented a set of classification techniques using the primary dataset composed of oil anomaly abstracts, and two auxiliary datasets from PubMed and google news. The tested classifiers designed to identify the text subject from the anomaly classes presented about 70 \% accuracy for the Logistic Regression and SVM classifier using the TF-IDF tokenizer, and for the Bert classifier using only the Nouns present in the abstracts. These results confirmed some suspects in topic modeling: some of the texts talk about more the one anomaly (especially anomalies 6 and 7), and there is some degree of intersection between the subjects. Although the Bert classifier is slightly less accurate than the other two, the generalization due to the sizeable pre-trained model is very compelling for method generalization.

The text segmentation classifier presented a good correspondence to the training and validation set, and can also be applied with a slight loss of accuracy in our primary dataset. This allows the information extraction for the methods and correspondent results for general texts. The combination with Bert network's sentiment analysis allows ranking the methods and results based on the class distance score. 

Combining the methods presented in this work corresponds to a practical strategy to analyze text abstracts. Despite the validation set composed of oil production problems, the strategy can evaluate different problems and processes.

\bibliographystyle{unsrt}  
\bibliography{references}

\begin{thebibliography}{10}

\bibitem{VARGAS2019}
Ricardo Emanuel~Vaz Vargas, Celso~Jos\'e Munaro, Patrick~Marques Ciarelli,
  Andr\'e~Gon\'calves Medeiros, Bruno~Guberfain do~Amaral, Daniel~Centurion
  Barrionuevo, Jean Carlos~Dias de~Ara\'ujo, Jorge~Lins Ribeiro, and Lucas
  Pierezan~Magalh\ aes.
\newblock A realistic and public dataset with rare undesirable real events in
  oil wells.
\newblock {\em Journal of Petroleum Science and Engineering}, 181:106223, 2019.

\bibitem{bert}
Jacob Devlin, MingWei Chang, Kenton Lee, and Kristina Toutanova.
\newblock Bert: Pre-training of deep bidirectional transformers for language
  understanding.
\newblock {\em CoRR}, abs/1810.04805, 2018.

\bibitem{romulo2019}
R{\^o}mulo C{\'e}sar~Costa de~Sousa and H{\'e}lio Lopes.
\newblock Portuguese pos tagging using blstm without handcrafted features.
\newblock In Ingela Nystr{\"o}m, Yanio Hern{\'a}ndez~Heredia, and Vladimir
  Mili{\'a}n~N{\'u}{\~{n}}ez, editors, {\em Progress in Pattern Recognition,
  Image Analysis, Computer Vision, and Applications}, pages 120--130, Cham,
  2019. Springer International Publishing.

\bibitem{Marulli2019}
Fiammetta Marulli, Marco Pota, and Massimo Esposito.
\newblock {\em A Comparison of Character and Word Embeddings in Bidirectional
  LSTMs for POS Tagging in Italian}, pages 14--23.
\newblock 01 2019.

\bibitem{Angelov}
Dimo Angelov.
\newblock Top2vec: Distributed representations of topics, 08 2020.

\bibitem{Madrid2019}
Jorge~G. Madrid and Hugo~Jair Escalante.
\newblock Meta-learning of text classification tasks.
\newblock In Ingela Nystr{\"o}m, Yanio Hern{\'a}ndez~Heredia, and Vladimir
  Mili{\'a}n~N{\'u}{\~{n}}ez, editors, {\em Progress in Pattern Recognition,
  Image Analysis, Computer Vision, and Applications}, pages 107--119, Cham,
  2019. Springer International Publishing.

\bibitem{huang2015bidirectional}
Zhiheng Huang, Wei Xu, and Kai Yu.
\newblock Bidirectional lstm-crf models for sequence tagging, 2015.

\bibitem{Dernoncourt}
Franck {Dernoncourt} and Ji~Young {Lee}.
\newblock {PubMed 200k RCT: a Dataset for Sequential Sentence Classification in
  Medical Abstracts}.
\newblock {\em arXiv e-prints}, page arXiv:1710.06071, October 2017.

\bibitem{Peralta2019}
Billy Peralta, Victor Tirapegui, Christian Pieringer, and Luis Caro.
\newblock A simple proposal for sentiment analysis on movies reviews with
  hidden markov models.
\newblock In Ingela Nystr{\"o}m, Yanio Hern{\'a}ndez~Heredia, and Vladimir
  Mili{\'a}n~N{\'u}{\~{n}}ez, editors, {\em Progress in Pattern Recognition,
  Image Analysis, Computer Vision, and Applications}, pages 152--162, Cham,
  2019. Springer International Publishing.

\bibitem{Shinha2020}
Ankur Shinha.
\newblock {Sentiment Analysis for Financial News}, 2020.

\bibitem{cai2018}
Shuang Cai, Ahmet Palazoglu, Laibin Zhang, and Jinqiu Hu.
\newblock Process alarm prediction using deep learning and word embedding
  methods.
\newblock {\em ISA Transactions}, 85, 10 2018.

\bibitem{lopes2020}
Lucas Lopes.
\newblock {PNL analysis in oil data}, 2020.

\end{thebibliography}

\end{document}